# A Multi-Probe Audit of Clinical-Interview Depression Detection Benchmarks

Takehiro Ishikawa[1,*] and Jon Duke[1,2]

[1] College of Computing, Georgia Institute of Technology, Atlanta, GA, USA
[2] Georgia Tech Research Institute, Georgia Institute of Technology, Atlanta, GA, USA

[*] Corresponding author: tishikawa8@gatech.edu

## Abstract

This paper audits benchmark evaluation in clinical-interview depression detection through four complementary probes across DAIC/E-DAIC, CMDC, ANDROIDS, MODMA, and PDCH. First, we re-evaluate E-DAIC under strict subject-disjoint leave-one-subject-out cross-validation. A lightweight hybrid text-plus-LLM-score model reaches macro-F1 = 0.723 — the highest reported under this protocol, to our knowledge — providing a conservative out-of-fold reference point that does not depend on the privileged official holdout. Second, we test whether the E-DAIC official split supports fine-grained leaderboard rankings by sweeping 96 model configurations across modality bundles, pooling strategies, and learners. Development-side cross-validation and official-test rankings align only moderately: the best cross-validation configuration ranks twentieth on the official test, the official-test winner ranks forty-first by cross-validation, top-3 overlap is zero, and the apparent winner is rank-1 in only 32.3% of subject bootstraps. Third, we externally validate strong public CMDC and ANDROIDS baselines that achieve near-ceiling in-domain performance. Zero-shot transfer to external corpora is substantially weaker. Finally, we stress-test E-DAIC text and audio models using paired symptom-dense versus symptom-light interview slices defined by an SRDS-based annotator. Text scores rise sharply on symptom-dense slices, whereas audio scores remain nearly flat; the text-minus-audio gap is positive across all five seeds.

## 1. Introduction

Depression is a major global public-health and clinical burden. The World Health Organization estimates that approximately 332 million people worldwide have depression, including 5.7% of adults, and depression can impair relationships, school and work functioning, and increase suicide risk [1]. Consistent with this scale, the Global Burden of Disease 2021 analysis estimated 332.41 million prevalent cases and 56.33 million disability-adjusted life years attributable to depressive disorders in 2021 [2].

The research community has increasingly sought objective and passively collectable biomarkers that can support earlier risk stratification before a full clinical picture is manifest [3], [4]. In this context, interview-derived audio, visual, and language signals are especially attractive because they can be captured with relatively low additional burden during clinical, semi-structured, or telehealth-style interactions, and because they offer complementary views of depressive presentation through vocal behavior, facial or behavioral cues, and linguistic content [5]-[7].

The DAIC family is among the most heavily reused benchmark families in depression prediction research [5], [6]. A 2026 systematic review of language-based depression detection found that 73 of 131 model-performance estimates used DAIC or a derivative dataset, and a 2024 speech meta-analysis found DAIC-WOZ was the most-used speech dataset among included studies [8], [9].

The dominant evaluation practice on E-DAIC relies on a single official split of 163 training, 56 development, and 56 test participants, and most studies report model performance on this single official test [6]. When such a small holdout is privileged, researchers can easily tune model selection, feature selection, thresholds, and reporting practices around it, creating a risk that headline scores appear higher than true performance [10]-[12]. Even without explicit misconduct, these choices can drift toward the official split, making split-specific wins look like general superiority. Compounding this, when the official test set is small, slight differences in macro-F1 — the field's standard metric — can arise from finite-sample noise or split-specific shifts rather than from genuine model superiority [13]-[15]. If leaderboard first-place finishes or differences of a few points are read directly as "Method A is superior to Method B," the entire field risks being pulled toward unstable leaderboard optimization rather than toward robust subject-level markers.

A second, related concern is participant-level leakage. In clinical interview data, splitting randomly at the turn, window, or question level leaks information from the same participant into both training and test sets, creating a risk that the model memorizes speaker idiosyncrasies or session-specific features rather than cues of depression. Recent work has warned about inflated or overly optimistic performance in DAIC-family and clinical-interview depression-detection settings when speaker or participant independence is not preserved [16]-[19]. In one such review of 66 studies entering quality assessment, at least six (9.1%) exhibited subject leakage, and another sixteen (24.2%) could not be verified because preprocessing and split construction were insufficiently documented; the same review experimentally showed that leakage substantially overestimated validation performance [16].

However, studies that report performance under subject-disjoint and out-of-fold cross-validation remain scarce [16]. These observations motivate the first two probes of this study. The aim of Probe A is to move away from the optimistic reading that relies on the official single split and to re-establish an anchor point for E-DAIC under subject-disjoint out-of-fold cross-evaluation that does not depend on the privileged official holdout. Probe B then asks the complementary question: even when the official split is used in good faith, is it stable enough to support the fine-grained model rankings that the field routinely draws from it?

A separate concern arises outside the DAIC family. On the CMDC and ANDROIDS datasets, near-ceiling baselines exist that achieve F1 / macro-F1 of approximately 0.95 in-domain under subject-disjoint cross-validation, and source code for these pipelines is publicly available [20]-[23]. The more a publicly available pipeline appears "solved," the more it needs auditing via zero-shot external evaluation before subsequent research inherits it as a settled baseline; without external validation, high in-domain scores are easily misread as indicating a solved problem.

Recent surveys and meta-analyses already point to a field dominated by a few reused corpora and limited generalizability evidence [8], [9], [24], [25], while cross-task or cross-language benchmark studies remain comparatively rare [26]. Probe C addresses this gap by systematically applying these strong, publicly available baselines to multiple external corpora.

Finally, within depression prediction the text modality has tended to yield the highest performance, compared to other modalities like audio and vision, and has therefore been treated as the strongest modality [27]-[31]. This raises the question of what the dominant text-based signal actually reflects. Probe D stress-tests E-DAIC text and audio against SRDS-defined symptom-dense content to ask what that dominant signal seems to be as clinical interviews, contain questions and topics directly related to depressive symptoms [32], [33].

The first two contributions of this paper concern evaluation rigor on E-DAIC. In Probe A, we construct a model whose T+L macro-F1 of 0.723 serves as a strong reference point under subject-disjoint and out-of-fold cross evaluation — the highest reported performance under this protocol, to our knowledge. Because the source code is publicly released [34], the community can build from this baseline, making it a reusable anchor for future work. At the same time, this figure is lower than optimistic reports based on the official split [16]-[19]. In other words, high scores on the single official split should not be read as stable performance across E-DAIC as a whole.

At the same time, Probe B shows that fine-grained rankings on the official split are weak evidence of stable model superiority. This matters because the DAIC family is heavily used in the field [8], [9] and comparing papers via official-test reports is the established practice; if the small official split is unstable, then comparing papers via fine-grained rankings on it is itself risky. In particular, differences of a few points or first-place claims should not be read too strongly unless uncertainty is shown via bootstrap or multiple splits [10], [12]-[15]. This probe does not claim that the official split is worthless; rather, it claims that reading a small, widely used holdout as the absolute arbiter of a leaderboard is dangerous.

The remaining two contributions look beyond E-DAIC's evaluation protocol. Probe C shows that even when source-side F1 / macro-F1 is approximately 0.95 [22], [23], this performance does not necessarily transfer to external data. The contribution here is not merely that external scores were low; rather, in a landscape where external-data evaluation is lacking [8], [9], [24], [26], strong publicly available baselines are systematically applied to multiple external corpora for the first time, turning the missing due diligence into a reusable reference point. Concretely, this prevents the research community from over-optimistically interpreting near-ceiling in-domain numbers as a solved baseline, and it provides — to our knowledge — the first systematic external evaluation of CMDC and ANDROIDS pipelines on multiple external corpora including MODMA and PDCH [35]-[37]. In Probe D, we provide evidence that the primary signal of E-DAIC text models is likely strongly dependent on SRDS-defined symptom-dense content. Because the text modality is the key modality in this research field [27]-[31], sharpening the interpretability of its high performance is critical.

## 2. Method

### 2.1 Datasets, labels, and audit scope

DAIC-WOZ and its extended E-DAIC variant are English interview datasets built around PHQ-8 style depression labels [5], [6]. DAIC-WOZ distributes two-speaker (interviewer + participant) transcripts, but E-DAIC provides only the participant side [32]. To supply question-response context for the LLM assessor, we separately generated a full-dialogue ASR transcript from the raw E-DAIC interview audio for each participant. Raw interview audio was converted to 16-kHz mono WAV, transcribed with faster-whisper large-v3, speaker-diarized with a two-speaker pyannote, and saved as chronological speaker/text/start/end utterance CSVs.

CMDC is a Chinese semi-structured interview corpus intended for preliminary screening [20]. MODMA is a public multimodal dataset for mental-disorder analysis [35], [36], and this paper acknowledges Gansu Provincial Key Laboratory of Wearable Computing, Lanzhou University, China as the source of the MODMA data. ANDROIDS provides a public speech-based depression benchmark from doctor-patient style dialogue [21]. PDCH adds clinically grounded consultation sessions with HAMD-17 assessments [37]. For PDCH, we evaluate three clinically meaningful binary label conventions, HAMD >= 8, >= 17, and >= 24, rather than assume a single cutoff.

These datasets differ in language, task framing, label definition, interview protocol, and metadata availability. Consequently, two papers can report superficially similar subject-level scores while relying on incompatible splits, different speaker filtering rules, or different assumptions about adaptation. Survey work has already pointed out that reproducibility and generalizability remain weak across this landscape [24].

Our analysis is subject-level throughout. Model outputs are aggregated to one decision per participant, and no participant appears in both train and test within a probe. This rule is intended to prevent row-, turn-, window-, question-, session-, or speaker-level subject leakage. For the zero-shot external audits, the target datasets are used only for final scoring; they are not used for target-label tuning, calibration fitting, or representation adaptation.

## 2.2 E-DAIC hybrid model

For Probe A, we evaluate E-DAIC under leave-one-subject-out cross-validation using a lightweight asymmetric hybrid network. We denote the main supervised configuration as T+L. The T branch uses E-DAIC transcript text. For each turn, we concatenate embeddings from sentence-transformers/all-mpnet-base-v2 and intfloat/e5-large-v2, yielding a 1792-dimensional text vector. Embeddings are extracted with maximum length 256, batch size 64, and L2 normalization. We also compute four turn-structure features for each participant turn: log(1 + turn duration), log(1 + previous gap), log(1 + character length), and log(1 + word length).

The L-only input is a participant-level probability generated by a separate openai/gpt-oss-120b inference notebook. For each participant, that notebook reads a full-dialogue ASR transcript, concatenates the utterance text lines, truncates the resulting document to at most 12,000 tokenizer tokens with a head-tail policy, and sends a two-message chat prompt to the model. The system prompt is reproduced verbatim below:

*"The following is a conversation between a participant and an interviewer. Take on the role of an expert in psychiatric diagnosis using the DSM-5. Estimate PHQ-8 item scores (0-3 each), compute total (0-24), and output depression_prob (0-1) and depression_label (1 if total>=10, else 0). Output valid JSON only: {"phq8_items": [int,...], "phq8_total": int, "depression_prob": float, "depression_label": int, "evidence_summary": string}"*

The user message is simply *Transcript:\n{doc}.* Decoding is deterministic (do_sample=False, max_new_tokens=2048, pad_token_id=eos), so no temperature, top-p, or seed is used. The scalar consumed by T+L is the returned depression_prob, clamped to [0,1]. For L-only row in Probe A, this participant-level probability is converted to a binary decision by the nested LOSO decision rule described in Section 2.6. This participant-level scalar is copied to every turn for that participant as X_l, i.e., the same scalar feature.

The hybrid model does not train or fine-tune the LLM-score generator, and no fold-specific fitting is performed on outer-test participants; the supervised model only consumes the resulting scalar as a precomputed input. These T+L and standalone L-only implementations for the Probe A E-DAIC LOSO runs are archived in the Zenodo v2 release [34].

Architecturally, the main branch adapts text, LLM, and structure features into a 192-dimensional latent sequence, applies FiLM-style gating, and feeds the fused representation to a bidirectional GRU with hidden size 192 and attention dimension 128. Audio and vision pass through smaller gated residual adapters (64 and 48 latent dimensions). We train with batch size 8, learning rate 2e-4, dropout 0.20, eight epochs, and an auxiliary turn-level loss weight of 0.20. In the strict LOSO run, every outer test participant is held out completely, and inner model selection uses only the remaining subjects.

Including interviewer utterances as text embeddings in a supervised text classifier's training can create shortcuts whereby the classifier exploits question identity, question order, or specific segments that ask about mental health history as proxy features for the label, rather than learning the participant's linguistic characteristics. This concern is especially relevant when interviewer text is fed into the classifier as high-dimensional, trainable supervised features [33]. However, the L modality in our study differs from this setting. The LLM score generator is neither trained nor fine-tuned on E-DAIC labels, folds, or split membership; instead, it reads the chronological transcript as a fixed full-dialogue assessor grounded in DSM-5/PHQ-8 criteria.

Furthermore, the only value passed to the downstream LOSO classifier is a one-dimensional scalar, depression_prob, at the participant level; the vocabulary, identity, order, and position of interviewer prompts are never provided as trainable text features. Consequently, the supervised prompt-identity/position shortcut pathway problematized by Burdisso et al. [33] does not exist in our design.

### 2.3 CMDC text baseline

For zero-shot external validation from Chinese interview text, we reproduce the public Xia question-level pipeline [22] built around nghuyong/ernie-3.0-base-zh embeddings and a LogisticRegression classifier (max_iter = 1000, random_state = 42). Each question-answer unit is embedded independently, question probabilities are averaged to a subject score, and the released source score-to-label rule is carried unchanged for zero-shot scoring. Under Xia's own participant-wise 5-fold StratifiedGroupKFold protocol, the reported cross-validated performance is approximately accuracy approx. 0.98, F1 approx. 0.96, and AUC approx. 0.97; our faithful reproduction of that same protocol yields participant-level F1 approx. 0.950 and AUROC approx. 0.980. We also apply a length-safe patch for external evaluation: overlength units are split by token budget rather than silently truncated, subchunk probabilities are averaged back to the original unit, and subject scores are then computed exactly as in the source pipeline.

### 2.4 ANDROIDS pipeline

We reproduce the public Daly and Olukoya ANDROIDS interview-task pipeline [23] with separate audio and text branches. The audio branch uses 32-dimensional numeric segment features, a 1-D convolution, max pooling, a 2-layer LSTM (hidden size 128), and attention-based segment aggregation. The text branch fine-tunes neuraly/bert-base-italian-cased-sentiment over transcript segments (max length 512; batch size 8). Subject-level fusion follows the released implementation's branch weighting. Daly and Olukoya report decision-level fusion accuracy = 94.30% and F-Score = 94.51%; using the released prediction files for the interview-task subset audited here, we exactly reconstruct the source-side 5-fold participant-level fusion result (mean macro-F1 = 0.955 and mean accuracy = 0.958).

*Table 1. Overview of the four-probe benchmark-audit protocol used in this study.*

| Probe | Question | Protocol | Primary outputs |
|---|---|---|---|
| A | Subject-disjoint out-of-fold cross-validated performance on E-DAIC | LOSO | Macro-F1, AUROC, AP |
| B | Official-split instability on E-DAIC | 96 configs = 8 modality bundles x 2 poolers x 6 learners; 3 seeds; 5-fold CV on 219 non-test subjects; official-test bootstrap | Rank overlap, regret, p(rank-1) |
| C | Zero-shot external validation | CMDC text baseline under zero-shot scoring on primary language-matched Chinese targets (MODMA, PDCH) plus supplementary cross-lingual checks; reproduced ANDROIDS pipeline audited separately by modality | Macro-F1, AUROC |
| D | Whether text is more sensitive than audio to SRDS-defined symptom-dense content | Heavy-topic versus neutral-topic pairs on E-DAIC, audio versus text, pre-specified SRDS slice builder, 5 seeds | Mean heavy-minus-neutral shift, paired gap, permutation p |

### 2.5 Concrete search spaces and slice construction

The official-split instability sweep is deliberately broad but computationally lightweight. It uses eight modality bundles—audio, vision, text, their A/V/T combinations, and A+V+T+L—two turn-to-subject poolers, and six tabular learners, yielding 8 x 2 x 6 = 96 unique configurations. Here, L denotes the one-dimensional LLM transcript score. Every configuration is run with the same three seeds and the same 5-fold stratified development-side CV protocol. Table 2 specifies the search space.

*Table 2. Concrete design of the 96-configuration official-split sweep.*

| Axis | Values | Count |
| --- | --- | --- |
| Modality bundles (8) | A; V; T; A+V; T+A; T+V; T+A+V; A+V+T+L | 8 |
| Poolers | mean; meanstd | 2 |
| Learners | logreg_pca64_c01; logreg_pca128_c1; linsvc_pca128_c1; extratrees_400; histgb_256; mlp_pca64_128 | 6 |
| Evaluation and uncertainty | 5-fold stratified CV on the 219 non-test subjects; seeds = {13, 23, 37}; official-test probabilities averaged over folds then seeds; 4000 subject bootstraps | 96 x 3 runs |

For the topic-sensitivity stress test, we construct paired heavy-topic and neutral-topic slices from E-DAIC participant content. Slice extraction is produced by a pre-specified automated topic-lite SRDS annotator rather than by post-hoc manual labeling: each chunk is scored with a small rubric that returns topic_score {0,1,2,3}, self_relevance {self, other, generic, negated}, and confidence, using local gpt-oss-20b inference with left/right context only for reference resolution.

The system prompt is reproduced verbatim below:

*"You are annotating chunk-level depression-topic density for research. This is not diagnosis. Score ONLY TARGET_CHUNK. Use LEFT_CONTEXT and RIGHT_CONTEXT only to resolve speaker reference, self-vs-other, or negation. Return only compact JSON, with no prose, no markdown, and no extra keys. Fields: topic_score: integer 0|1|2|3 (0 = no depression-related topic or cue; 1 = weak / ambiguous / indirect depressive cue or low-density mention; 2 = clear depression-related topic or likely self-relevant depressive cue; 3 = strong, direct, repeated, or highly salient self-relevant depressive topic/cue); self_relevance: one of self|other|generic|negated; confidence: float between 0 and 1."*

The user message supplies three labelled fields—TARGET_CHUNK, LEFT_CONTEXT, and RIGHT_CONTEXT—followed by the instruction "Return JSON only." Decoding is deterministic (do_sample=False). Heavy slices correspond to SRDS-defined symptom-dense windows; neutral slices correspond to SRDS-defined symptom-light windows. Rows assigned to the bookkeeping-only topic_mid category are excluded rather than interpreted as a substantive middle condition. Each band must contain at least 10 seconds of participant speech, and only participants with both bands are retained. The final paired evaluation therefore uses 275 participant pairs for text and 132 participant pairs on the modality-overlap subset used for the paired text-minus-audio comparison.

This choice is deliberate because one pre-specified automated annotator reduces researcher degrees of freedom relative to post-hoc manual slicing, even though the resulting slice boundaries remain rubric-dependent. We do not claim that SRDS recovers ground-truth clinical topic density; the inferential target is narrower: under a transparent and pre-specified instrument, are text models unusually sensitive to SRDS-defined symptom-dense content relative to audio? The paired within-participant design then isolates that sensitivity from between-subject differences, while causal or confounding interpretations remain secondary to the stress-test framing.

### 2.6 Statistical reporting, decision rules, and sensitivity analysis

Our primary metric is participant-level macro-F1, which is commonly reported in binary clinical-interview depression-detection evaluations and remains interpretable under class imbalance. We also report AUROC and, where relevant, average precision. For Probes B and C, score-to-label conversion uses the pre-specified source/default rule y-hat = 1[s(x) >= 0.5]. For L-only in Probe A, the binary decision is instead produced by a nested participant-wise LOSO rule that maximizes macro-F1 on the outer-training subjects; ties are broken by choosing the cutoff closest to 0.5.

For probe B, we report Pearson, Spearman, and Kendall associations between cross-validation and official-test scores, rank overlap at multiple cutoffs, and the distribution of bootstrap ranks on the official test. The bootstrap resamples official-test subjects with replacement 4000 times.

For probe D, let Delta_i = p_i(heavy) - p_i(neutral) for participant i. We estimate the mean shift with 5000 paired participant-level bootstrap replicates. To test whether text moves more than audio on the same participants, we compute the paired gap Delta_i(text) - Delta_i(audio) and apply a sign-flip permutation test with 5000 random sign assignments. The reduced multi-seed analysis uses five seeds: 13, 23, 37, 42, and 79.

## 3. Results

### 3.1 Subject-disjoint out-of-fold cross-validated performance on E-DAIC under LOSO

Table 3 contrasts the performance of L-only, T-only, and T+L evaluated on E-DAIC under LOSO. L-only probability ranks subjects strongly (AUROC = 0.825, AP = 0.666). When the decision cutoff is selected inside each LOSO fold on the training participants to maximize macro-F1, the pooled out-of-fold L-only result reaches macro-F1 = 0.686, accuracy = 0.713, and balanced accuracy = 0.702, with TN = 138, FP = 51, FN = 28, and TP = 58. The selected cutoffs are highly stable across folds (mean = 0.150, sd = 0.004; median = 0.15; range = 0.13-0.17). T-only reaches macro-F1 = 0.621, AUROC = 0.647, and AP = 0.438, with confusion counts TN = 141, FP = 48, FN = 43, and TP = 43.

T+L improves to macro-F1 = 0.723, AUROC = 0.768, and AP = 0.592, with confusion counts TN = 155, FP = 34, FN = 32, and TP = 54. This score is lower than many optimistic official-split claims, but it is subject-disjoint and strictly out-of-fold at the participant level; the evaluation does not depend on a privileged official holdout. On a benchmark family heavily reused by the field [8], [9], resetting that anchor matters. Additional audio and vision checks did not consistently outperform the T+L core.

*Table 3. Subject-disjoint out-of-fold cross-validated performance on E-DAIC under LOSO.*

| Config | Macro-F1 | AUROC | AP | TN / FP / FN / TP |
|---|---|---|---|---|
| L-only | 0.686 | 0.825 | 0.666 | 138 / 51 / 28 / 58 |
| T-only | 0.621 | 0.647 | 0.438 | 141 / 48 / 43 / 43 |
| T+L | 0.723 | 0.768 | 0.592 | 155 / 34 / 32 / 54 |

### 3.2 Official-split ranking instability on E-DAIC

Table 4 summarizes the 96-configuration official-split audit. Even after averaging over three seeds, cross-validation and official-test performance align only moderately. The best cross-validation configuration is only twentieth on the official test, whereas the best official-test configuration is forty-first by cross-validation. The top-3 overlap is zero, the top-5 overlap is one, and the median absolute rank shift is 15.5 positions.

The subject bootstrap sharpens the point. Even the test-best configuration has a bootstrap probability of 0.323 of being truly rank-1, and its 95 percent rank range spans roughly positions 1 to 19. This means that a narrow top-of-table win on the official split is not a strong basis for claiming stable superiority. We do not attribute this instability solely to random folds, because the official split also contains train-test shift; rather, the results show that fine-grained leaderboard ordering on this small official split is unsafe to read literally.

*Table 4. Summary of the 96-configuration official-split instability analysis.*

| Statistic | Value | Reading |
|---|---|---|
| Pearson correlation (CV vs official test) | 0.6373 | Moderate linear alignment across 96 configs. |
| Spearman correlation (CV vs official test) | 0.7037 | Rank order aligns only partially. |
| Kendall tau | 0.4884 | Pairwise ordering disagreement remains substantial. |
| Discordance rate | 0.2545 | About 25.4 percent of pairwise comparisons reverse. |
| Best-CV config test rank | 20 | The dev-side winner is not the official-test winner. |
| Best-test config CV rank | 41 | The official-test winner looks mediocre by CV. |
| Top-3 overlap | 0 | No shared systems between the two top-threes. |
| Top-5 overlap | 1 | Only one system is shared between the two top-fives. |
| Median absolute rank shift | 15.5 | Typical top-line movement is large. |
| Bootstrap p(rank-1) of test-best config | 0.323 | Even the apparent winner is first in only 32.3 percent of bootstraps. |
| Bootstrap 95% rank range of test-best config | 1-19 | Uncertainty still spans much of the leaderboard. |

### 3.3 External validation of the reproduced CMDC baseline

Table 5 summarizes zero-shot subject-level behavior for the CMDC text baseline. In the faithful Xia-style source reproduction [22], the original subject-disjoint 5-fold out-of-fold protocol yields participant-level F1 approx. 0.950 and AUROC approx. 0.980, close to the source paper's reported CMDC accuracy approx. 0.98, F1 approx. 0.96, and AUC approx. 0.97.

Under zero-shot external evaluation, however, this near-ceiling source-side performance does not translate into strong external classification performance. Macro-F1 drops to 0.265 on MODMA and ranges from 0.127 to 0.446 across the three PDCH label cutoffs. On the language-matched targets, AUROC is 0.672 on MODMA and 0.564 on PDCH >= 8, but decreases to 0.442 and 0.422 under the stricter PDCH >= 17 and PDCH >= 24 label definitions, respectively. This pattern suggests that the reproduced CMDC text pipeline retains limited ranking signal in some external Chinese settings, but its source score-to-label rule does not provide robust zero-shot classification across target corpora or clinical cutoffs. The cross-lingual checks show similarly weak transfer. When applied to E-DAIC and ANDROIDS, the CMDC text baseline reaches macro-F1 = 0.238 and 0.420, respectively, with AUROC = 0.579 and 0.468.

*Table 5. External performance of the reproduced CMDC text baseline.*

| Dataset | N | Macro-F1 | AUROC |
|---|---|---|---|
| MODMA (primary) | 36 | 0.265 | 0.672 |
| PDCH >= 8 (primary) | 62 | 0.127 | 0.564 |
| PDCH >= 17 (primary) | 62 | 0.361 | 0.442 |
| PDCH >= 24 (primary) | 62 | 0.446 | 0.422 |
| E-DAIC (supp.) | 275 | 0.238 | 0.579 |
| ANDROIDS (supp.) | 116 | 0.420 | 0.468 |

### 3.4 External evaluation of the reproduced ANDROIDS pipeline

Table 6 summarizes the external audit of the public Daly and Olukoya ANDROIDS implementation [23]. Their article reports decision-level fusion accuracy = 94.30% and F-Score = 94.51%; using the released prediction files for the interview-task subset audited here, we exactly reconstruct the source-side subject-disjoint 5-fold out-of-fold participant-level fusion result (mean macro-F1 = 0.955 and mean accuracy = 0.958).

The source-side fusion advantage does not survive external evaluation. Across the audited targets, the audio branch is generally the most portable component: it gives the strongest macro-F1 and AUROC on CMDC, and the strongest AUROC on PDCH. In contrast, the fusion and text branches often perform substantially worse under zero-shot transfer. This is especially visible on CMDC, where audio reaches macro-F1 = 0.647 and AUROC = 0.768, while fusion and text both reach macro-F1 = 0.400, and the text branch has AUROC = 0.288.

The text-branch result should be interpreted cautiously. The released ANDROIDS text branch uses an Italian sentiment encoder, so poor transfer to Chinese or English corpora should not be read as a universal failure of text-based depression modeling. Rather, Table 6 shows that a near-ceiling source-side multimodal pipeline can rely on corpus-specific or language-specific regularities that do not automatically transfer. For this reason, we treat the ANDROIDS experiment as a reproducibility and external-audit probe, not as a claim that audio is intrinsically superior to text across depression-detection settings.

*Table 6. External subject-level performance of the reproduced ANDROIDS pipeline by modality.*

| Target | Modality | N | Macro-F1 | AUROC |
|---|---|---|---|---|
| CMDC | Audio | 78 | 0.647 | 0.768 |
| CMDC | Fusion | 78 | 0.400 | 0.630 |
| CMDC | Text | 78 | 0.400 | 0.288 |
| E-DAIC | Audio | 275 | 0.250 | 0.518 |
| E-DAIC | Fusion | 275 | 0.238 | 0.524 |
| E-DAIC | Text | 274 | 0.243 | 0.553 |
| MODMA | Audio | 36 | 0.390 | 0.482 |
| MODMA | Fusion | 36 | 0.265 | 0.425 |
| MODMA | Text | 36 | 0.265 | 0.378 |
| PDCH | Audio | 62 | 0.472 | 0.625 |
| PDCH | Fusion | 62 | 0.503 | 0.520 |
| PDCH | Text | 62 | 0.424 | 0.468 |

### 3.5 SRDS-defined symptom-density stress test

Figure 1 and Table 7 report the paired SRDS-conditioned stress-test results. The crucial design choice is that heavy-topic (i.e., SRDS-defined symptom-dense) and neutral (SRDS-defined symptom-light) slices come from the same participants, and slice construction is delegated to a pre-specified automated SRDS annotator rather than to post-hoc manual labeling.

Under that control, text scores rise sharply on symptom-dense slices, whereas audio scores move very little. With full data, the mean symptom-dense-minus-symptom-light shift is 0.422 for text, while audio remains approximately flat (mean -0.004).

The reduced multi-seed rerun makes Probe D harder to dismiss as a single-run artifact. Across seeds {13, 23, 37, 42, 79}, text shift is positive in 5/5 seeds, and its participant-bootstrap confidence interval excludes 0 in every seed. Audio remains near zero, and its confidence interval crosses 0 in all five seeds. On the 132-participant modality-overlap subset, the paired text-minus-audio gap is large and stable: mean 0.409 (sd 0.046), positive in all five seeds, with sign-flip permutation p = 0.0002 in every seed.

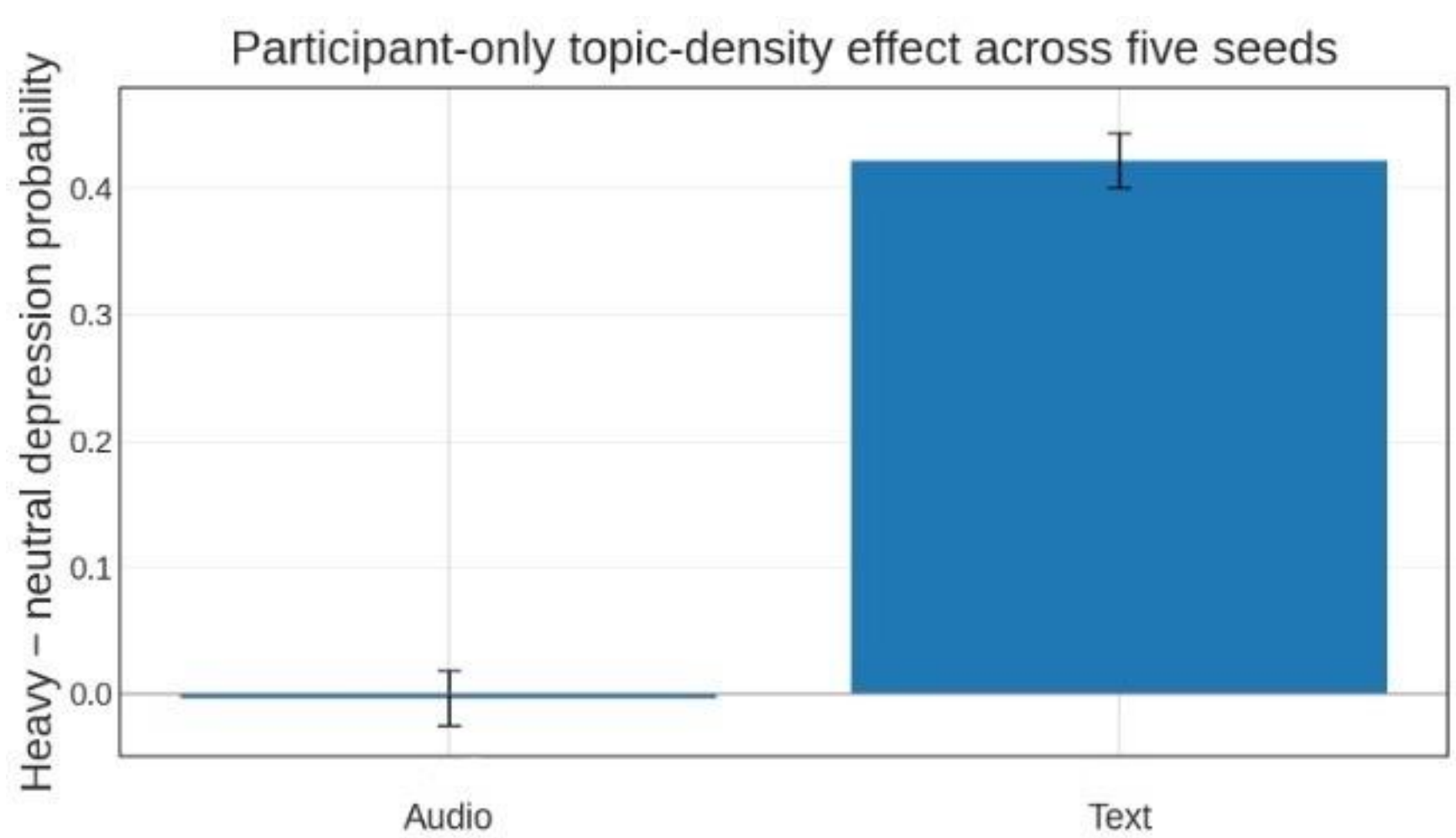

*Figure 1. Mean symptom-dense-minus-symptom-light probability shift for audio and text models in the paired same-participant SRDS-defined symptom-density stress test on E-DAIC slices.*

*Table 7. Mean symptom-dense-minus-symptom-light shift and text-minus-audio gap across five seeds for the same-participant SRDS-defined symptom-density stress test. Reported values are mean shifts with seed-level standard deviation in parentheses.*

| Audio shift | Text shift | Text - audio gap | Multi-seed summary |
|---|---|---|---|
| -0.004 (sd 0.022) | 0.422 (sd 0.022) | 0.409 (sd 0.046) | Text > 0 in 5/5 seeds; audio CI crosses 0 in all 5; gap > 0 in 5/5; p = 0.0002 each seed |

## 4. Discussion

On E-DAIC, Probe A provides a subject-disjoint and out-of-fold reference point that avoids relying on the privileged official test set. The T+L configuration reached macro-F1 = 0.723 under LOSO evaluation. The result is deliberately more conservative than many official-split-oriented reports; nevertheless, it is, to our knowledge, the highest reported performance under this protocol. Its value is therefore not only as a performance claim, but as a more conservative anchor for future work in the research community. In a field where DAIC-family datasets are repeatedly reused [8], [9], a reproducible LOSO reference point helps shift attention away from single-holdout maximization and toward subject-level robustness.

Probe B reinforces this point by showing that the E-DAIC official split is not stable enough to support fine-grained leaderboard conclusions. Cross-validation and official-test scores were only moderately aligned, the best development-side configuration ranked only twentieth on the official test, and the official-test winner ranked forty-first by development-side cross-validation. The top-3 overlap was zero, and even the apparent official-test winner was rank-1 in only 32.3% of bootstrap resamples. The results argue against treating small differences on that split as decisive evidence of model superiority. In this setting, claims such as “state of the art” or “best-performing method” should be accompanied by uncertainty estimates, bootstrap rank intervals, multiple subject-disjoint splits, or external validation.

The external audits in Probe C show that strong in-domain performance does not necessarily transfer beyond the source corpus. The reproduced CMDC and ANDROIDS pipelines achieved or reconstructed near-ceiling source-side performance, but their zero-shot external behavior was much weaker. This gap is important because high in-domain scores can easily create the impression that a benchmark has been solved. The present results suggest a different conclusion: these pipelines may be highly effective at exploiting source-corpus regularities, but those regularities do not automatically generalize across datasets. Differences in language, interview framing, clinical label definition, class balance, recording environment, participant population, and preprocessing conventions can all affect transfer. The contribution here is not merely that external scores were low; rather, in a landscape where external-data evaluation is lacking [8], [9], [24], [26], strong publicly available baselines are systematically applied to multiple external corpora for the first time, turning the missing due diligence into a reusable reference point. Concretely, this prevents the research community from over-optimistically interpreting near-ceiling in-domain numbers as a solved baseline, and it provides — to our knowledge — the first systematic external evaluation of CMDC and ANDROIDS pipelines on multiple external corpora including MODMA and PDCH.

Probe D provides a complementary explanation for why text often appears to be the strongest modality in depression-detection benchmarks. Under a paired same-participant design, text scores rose sharply on SRDS-defined symptom-dense slices, whereas audio scores remained close to flat. The text-minus-audio gap was positive across all five seeds and significant under sign-flip permutation testing. This suggests that the dominant text signal in E-DAIC is strongly tied to explicit symptom-related content. Such a signal is not necessarily invalid. In a clinical interview, direct discussion of mood, sleep, appetite, anhedonia, suicidality, or functional impairment is clinically meaningful. However, the interpretation differs from a claim that the model has discovered subtle, context-independent linguistic markers of depression. A text model may perform well because the interview protocol elicits symptom-relevant content, because certain questions are more diagnostically informative, or because depressed and non-depressed participants encounter or respond to different topic contexts. Given the text modality is the key modality in this research field [27]-[31], sharpening the interpretability of its high performance via probe D is critical.

## 5. Limitations

Regarding the limitation of possible pretraining exposure, DAIC/E-DAIC are controlled-access research corpora rather than open-web benchmarks, which makes lawful exact inclusion of their raw transcripts or labels in web-scale LLM training data substantially less plausible than for fully public NLP datasets. Nevertheless, because we cannot audit the LLM's complete pretraining corpus, we do not claim to certify that pretraining contamination is absent.

Because CMDC->E-DAIC and CMDC->ANDROIDS are cross-lingual as well as cross-corpus, we treat them as supplementary checks. The core external claim is already established by the language-matched Chinese transfers to MODMA and PDCH. The ANDROIDS text branch uses an Italian sentiment encoder, so its external text failure should not be generalized to all text models. For the small external targets, we also treat AUROC-versus-score-to-label readings as descriptive only; the paper's main external claim is the value of zero-shot external auditing, not a definitive decomposition of why each target fails.

The topic-sensitivity probe relies on SRDS-based slice construction rather than manual expert annotation. This is a deliberate design choice, not a post-hoc omission. A pre-specified automated annotator reduces researcher degrees of freedom, applies one rubric consistently to every participant, and avoids relabeling after inspecting downstream predictions. Our claim is therefore not that SRDS recovers ground-truth clinical topic density, but that text models are highly sensitive to SRDS-defined symptom-dense content under a transparent and reproducible instrument. The reduced multi-seed rerun lowers the risk that the reported direction is a single training-seed artifact, but it does not remove the fact that the stress test remains conditional on this specific SRDS rubric. That conditionality is a scope condition on interpretation, not an argument that the pre-specified automated annotator is weaker than post-hoc manual slicing. The exact magnitude of the effect is therefore extractor-dependent, and alternative rubrics could shift the absolute numbers.

## Data and code availability

All analyses use previously released or controlled-access research corpora, so readers should obtain dataset files from the original data providers under the applicable data-use conditions. The T+L LOSO implementation and L-only inference/decision-rule code used for Probe A are archived on Zenodo as Version v2, published Apr. 27, 2026 [34].

## Ethics statement

This study used previously released or controlled-access, de-identified research datasets and did not involve new recruitment, intervention, or direct contact with human participants by the authors.

## Declaration of AI use

AI-assisted tools were also used for language editing, manuscript polishing, and programming assistance, including drafting, refactoring, and debugging analysis code. All code, analyses, outputs, interpretations, claims, and final manuscript content were reviewed and approved by the human authors.

## Authors' contributions

T.I.: conceptualization, methodology, software, formal analysis, investigation, data curation, validation, visualization, writing – original draft, and writing – review and editing.

J.D.: supervision

## Competing interests

We declare we have no competing interests.

## Funding

This research received no external funding.

## Acknowledgements

The authors have no additional acknowledgements.